\documentclass{article}

\usepackage{arxiv}

\usepackage{natbib} 
\usepackage[utf8]{inputenc} 
\usepackage[T1]{fontenc}    
\usepackage{hyperref}       
\usepackage{url}            
\usepackage{booktabs}       
\usepackage{amsfonts,amsmath,amssymb}       
\usepackage{nicefrac}       
\usepackage{microtype}      
\usepackage{lipsum}
\usepackage{epsfig}
\usepackage{subfig}
\usepackage{setspace}

\usepackage{graphicx}
\usepackage{multirow}
\usepackage{todonotes}
\usepackage{stmaryrd,nicefrac}

\renewcommand{\vec}[1]{\boldsymbol{#1}}
\newcommand{\given}{\, | \,}
\newcommand{\hath}{\hat{h}}

\newcommand{\haty}{\hat{y}}

\newcommand{\fromto}{\longrightarrow}

\newcommand*{\defeq}{=}

\newcommand{\cX}{\mathcal{X}}
\newcommand{\cY}{\mathcal{Y}}
\newcommand{\cH}{\mathcal{H}}
\newcommand{\cD}{\mathcal{D}}
\newcommand{\eu}{\operatorname{EU}}
\newcommand{\au}{\operatorname{AU}}
\newcommand{\tu}{\operatorname{U}}
\newcommand{\Prob}{P}
\newcommand{\prob}{p}
\newcommand{\argmin}{\operatorname*{argmin}}

\newcommand{\on}[1]{\operatorname{#1}}

\begin{document}
\title{Ensemble-based Uncertainty Quantification: Bayesian versus Credal Inference}
%
%

\author{Mohammad Hossein Shaker\\
Department of Computer Science\\
Paderborn University\\
Paderborn, Germany\\
\texttt{mhshaker@mail.upb.de}\\
  \And
Eyke H{\"u}llermeier\\
Institute of Informatics\\
University of Munich (LMU)\\
Munich, Germany\\
\texttt{eyke@lmu.de}}

\maketitle

\begin{abstract}
The idea to distinguish and quantify two important types of uncertainty, often referred to as aleatoric and epistemic, has received increasing attention in machine learning research in the last couple of years. In this paper, we consider ensemble-based approaches to uncertainty quantification. Distinguishing between different types of uncertainty-aware learning algorithms, we specifically focus on Bayesian methods and approaches based on so-called credal sets, which naturally suggest themselves from an ensemble learning point of view. For both approaches, we address the question of how to quantify aleatoric and epistemic uncertainty. The effectiveness of corresponding measures is evaluated and compared in an empirical study on classification with a reject option. 

\end{abstract}

\section{Introduction}


In the literature on uncertainty, two inherently different sources of uncertainty are commonly distinguished, referred to as \emph{aleatoric} and \emph{epistemic} \citep{hora_aa96}. Broadly speaking, aleatoric (\emph{aka} statistical) uncertainty refers to the notion of randomness\,---\,coin flipping is a
prototypical example. 
As opposed to this, epistemic (\emph{aka} systematic) uncertainty refers to uncertainty caused by a lack of knowledge, i.e., it relates to the epistemic state of an agent. This uncertainty can in principle be reduced on the basis of additional information. In other words, epistemic uncertainty refers to the \emph{reducible} part of the (total) uncertainty, whereas aleatoric uncertainty refers to the \emph{non-reducible} part. 

This distinction has also been adopted in the recent machine learning literature, where the ``agent'' is a learning algorithm \citep{mpub272}. A distinction between aleatoric and epistemic uncertainty appears to be specifically useful in deep learning \citep{kend_wu17}, for example, where neural networks have been shown to have very limited awareness of their own competence. Often, such networks make fatal mistakes despite pretending high confidence, or can easily be fooled by ``adversarial examples'' \citep{pape_dk18}.

In this paper, we consider ensemble-based approaches to uncertainty quantification, i.e., to derive meaningful measures of aleatoric and epistemic uncertainty in a prediction. In this regard, we propose a taxonomy of different types of uncertainty-aware learning algorithms: probabilistic agents, Bayesian agents, and Levi agents (Section 3). While Bayesian agents are based on probabilistic uncertainty representation as commonly used in Bayesian inference, Levi agents are based on credal sets, i.e., (convex) sets of probability distributions, which provide the basis for a robust extension of traditional Bayesian statistics. Both approaches appear to be quite natural from an ensemble learning point of view. We address the question of how to quantify aleatoric and epistemic uncertainty in a formal way (Section 4), both for Bayesian and Levi agents, and how to approximate such quantities empirically using ensemble techniques (Section 5). Moreover, we analyze the effectiveness of corresponding measures in an empirical study on classification with a reject option (Section 6).


\section{Types of Uncertainty}\label{sec:EpisAlea}

We consider a standard setting of supervised learning, in which a learner is given access to a set of (i.i.d.) training data $\mathcal{D} \defeq \{ (\vec{x}_i , y_i )\}_{i=1}^N \subset \mathcal{X} \times \mathcal{Y}$, where $\mathcal{X}$ is an instance space and $\mathcal{Y}$ the set of outcomes that can be associated with an instance. In particular, we focus on the classification scenario, where $\cY=\{y_1, \ldots, y_K\}$ consists of a finite set of class labels, with binary classification ($\cY = \{0,1\}$) as an important special case.

Suppose a \emph{hypothesis space} $\mathcal{H}$ to be given, where a hypothesis $h \in \cH$ is a mapping $\cX \fromto \mathbb{P}(\cY)$, with $\mathbb{P}(\cY)$ the class of probability measures on $\cY$. Thus, a hypothesis maps instances $\vec{x}\in\cX$ to probability distributions on outcomes. The goal of the learner is to induce a hypothesis $h^* \in \mathcal{H}$ with low risk (expected loss)
\begin{equation}
R(h) \defeq \int_{\cX \times \cY} \ell( h(\vec{x}) , y) \, d \, \Prob(\vec{x} , y) \enspace ,
\end{equation}
where $P$ is the (unknown) data-generating process (a probability measure on $\cX \times \cY$), and $\ell: \, \mathbb{P}(\cY) \times \mathcal{Y} \longrightarrow \mathbb{R}$ a loss function. 
The choice of a hypothesis is commonly guided by the empirical risk 
\begin{equation}
R_{emp}(h) \defeq  \frac{1}{N} \sum_{i=1}^N \ell(h(\vec{x}) , y) \enspace ,
\end{equation}
i.e., the performance of a hypothesis on the training data. However, since $R_{emp}(h)$ is only an estimation of the true risk $R(h)$, the empirical risk minimizer (or any other predictor)
\begin{equation}\label{eq:argerm}
\hath \defeq \argmin_{h \in \cH} R_{emp}(h)
\end{equation}
favored by the learner will normally not coincide with the true risk minimizer (Bayes predictor)
\begin{equation}\label{eq:bayespred}
h^* \defeq \argmin_{h \in \cH} R(h) \, .
\end{equation} 
Correspondingly, there remains uncertainty regarding $h^{*}$ as well as the approximation quality of $\hath$ (in the sense of its proximity to $h^*$) and its true risk $R(\hath)$.

Eventually, one is often interested in the \emph{predictive uncertainty}, i.e., the uncertainty related to the prediction $\haty_{q}$ for a concrete query instance $\vec{x}_{q} \in \cX$. 
 Indeed, estimating and quantifying uncertainty in a transductive way, in the sense of tailoring it to individual instances, is arguably important and practically more relevant than a kind of average accuracy or confidence, which is often reported in machine learning. 
%
%
%
%
As the prediction $\haty_{q}$ constitutes the end of a process that consists of different learning and approximation steps, all errors and uncertainties related to these steps may also contribute to the uncertainty: 
\begin{itemize}
\item Since the dependency between $\cX$ and $\cY$ is typically non-deterministic, the description of a new prediction problem in the form of an instance $\vec{x}_{q}$ gives rise to a conditional probability distribution
\begin{equation}\label{eq:ccp}
\prob( y \given \vec{x}_{q}) = \frac{\prob(\vec{x}_{q} , y)}{\prob(\vec{x}_q)} 
\end{equation}
on $\cY$, but it does normally not identify a single outcome $y$ in a unique way. Thus, even given full information in the form of the measure $\Prob$ (and its density $\prob$), uncertainty about the actual outcome $y$ remains. This uncertainty is of an \emph{aleatoric} nature. In some cases, the distribution (\ref{eq:ccp}) itself (called the predictive posterior distribution in Bayesian inference) might be delivered as a prediction. Yet, when having to commit to a point estimate, the best prediction (in the sense of minimizing the expected loss) is prescribed by the pointwise Bayes predictor $f^*$, which is defined by
\begin{equation}\label{eq:pointbayespred}
f^*(\vec{x}) \defeq \argmin_{\haty \in \cY} \int_\cY \ell(y , \haty) \, d \Prob( y \given \vec{x} )
\end{equation} 
for each $\vec{x} \in \cX$.

\item The Bayes predictor (\ref{eq:bayespred}) does not necessarily coincide with the pointwise Bayes predictor (\ref{eq:pointbayespred}). This discrepancy between $h^*$ and $f^*$ is connected to the uncertainty regarding the right type of model to be fit, and hence the choice of the hypothesis space $\cH$. We refer to this uncertainty as \emph{model uncertainty}. Thus, due to this uncertainty, $h^*(\vec{x}) = f^*(\vec{x})$ cannot be guaranteed. Likewise, in case the hypothesis $h^*$ delivers probabilistic predictions $\prob(y \given h^*, \vec{x})$ instead of point predictions, one cannot assure that $\prob( \cdot \given h^*, \vec{x}) = \prob( \cdot \given \vec{x})$.

\item The hypothesis $\hath$ produced by the learning algorithm, for example the empirical risk minimizer (\ref{eq:argerm}), is only an estimate of $h^*$, and the quality of this estimate strongly depends on the quality and the amount of training data. We refer to the discrepancy between $\hath$ and $h^*$, i.e., the uncertainty about how well the former approximates the latter, as \emph{approximation uncertainty}. 

\end{itemize}
As already said, aleatoric uncertainty is typically understood as uncertainty that is due to influences on the data-generating process that are inherently random, that is, due to the non-deterministic nature of the sought input/output dependency. This part of the uncertainty is irreducible, in the sense that the learner cannot get rid of it. Model uncertainty and approximation uncertainty, on the other hand, are subsumed under the notion of epistemic uncertainty, that is, uncertainty due to a lack of knowledge about the perfect predictor (\ref{eq:pointbayespred}). Obviously, this lack of knowledge will strongly depend on the underlying hypothesis space $\cH$ as well as the amount of data seen so far: The larger the number $N = | \mathcal{D}|$ of observations, the less ignorant the learner will be when having to make a new prediction. In the limit, when $N \rightarrow \infty$, a consistent learner will be able to identify $h^*$. Moreover, the ``larger'' the hypothesis space $\cH$, i.e., the weaker the prior knowledge about the sought dependency, the higher the epistemic uncertainty will be, and the more data will be needed to resolve this uncertainty.


\section{Uncertainty Representation}


Given a query $\vec{x}_q$ for which a prediction is sought, different learning methods proceed on the basis of different types of information. Depending on how the uncertainty is represented as a basis for prediction and decision making, we propose to distinguish three types of learning methods, which we call, respectively, probabilistic, Bayesian, and Levi agents. 


\begin{figure}
\begin{center}
\includegraphics[scale=0.4]{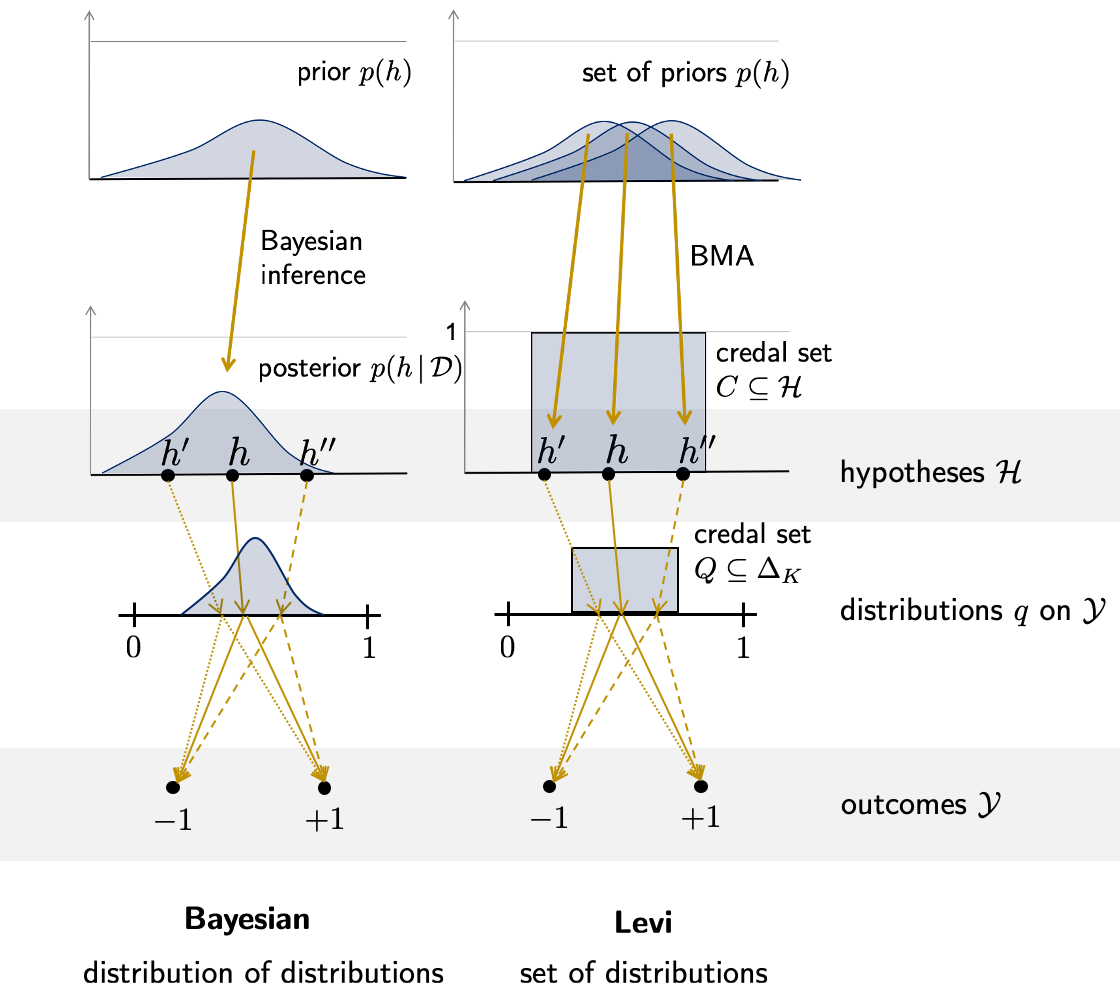} \qquad
\caption{Learning and prediction of a Bayesian compared to a Levi agent: Distribution of probability estimates versus set of expectation over such distributions.}
\label{fig:overview}
\end{center}
\end{figure}

\subsection{Probabilistic Agents}
A common practice in machine learning is to consider learners that fully commit to a single hypothesis $\hath \in \cH$ and use this hypothesis to make predictions. Such a learner will predict a single probability distribution 
\begin{equation}\label{eq:q}
q = \hath(\vec{x}_q) = (q_1, \ldots , q_K)   \in \mathbb{P}(\cY) \, ,
\end{equation}
where $q_k$ is the probability of the $k^{th}$ class $y_k$. This prediction is considered as an estimation of the conditional probability (\ref{eq:ccp}). 
We call a learner of that kind a \emph{probabilistic agent}. Such an agent's uncertainty about the outcome $y$ is purely aleatoric. At the level of the hypothesis space, the agent pretends full certainty, and hence the absence of any epistemic uncertainty.

\subsection{Bayesian Agents}
Adhering to the principle of (strict) Bayesianism as advocated by statisticians such as De Finetti \citep{defi_fi80}, 
a \emph{Bayesian agent} will represent its belief in terms of a probability distribution on $\cH$. Thus, instead of committing to a single hypothesis $\hat{h}$, the agent will assign a probability (density) $\prob(h)$ to each candidate $h \in \cH$. Moreover, belief revision in the light of observed data $\cD$ is accomplished by replacing this distribution with the posterior $\prob(h \given \cD)$.

Since every $h \in \cH$ gives rise to a probabilistic prediction (\ref{eq:q}), a Bayesian agent's belief about the outcome $y_q$ is represented by a second-order probability: a probability distribution of probability distributions. If needed, $\prob$ can be ``collapsed'' into a single distribution $q$ on $\cY$. This is typically accomplished by inducing $q$ from $\prob$ (or, more generally, a corresponding measure $\Prob$) via Bayesian model averaging (BMA):
\begin{equation}\label{eq:bma}
q \defeq \operatorname{bma}(\prob) = \int_{\cH} h(\vec{x}_q)  \, d \, \Prob(h) 
\end{equation}


\subsection{Levi Agents}

As a further generalization, instead of committing to a single probability distribution $\prob \in \mathbb{P}(\cH)$ on the hypothesis space, the learner may work with a \emph{set} $Q' \subseteq \mathbb{P}(\cH)$ of such distributions, 
all of which are deemed plausible candidates. 
Each distribution $\prob \in Q'$ again gives rise to a probability distribution according to (\ref{eq:bma}).
Eventually, the relevant representation of the learner is a set of probability distributions 
\begin{equation}\label{eq:credalset}
Q \defeq \big\{ \operatorname{bma}(\prob) \given \prob \in Q' \big\} \subseteq \mathbb{P}(\cY) \, .
\end{equation}
The reasonableness of taking decisions on the basis of sets of probability distributions (and thus deviating from strict Bayesianism) has been advocated by decision theorists like Levi \citep{levi_oi74,levi_te}. Correspondingly, we call a learner of this kind a \emph{Levi agent}.

The set $Q'$ (and thereby the set $Q$) can be produced in different ways, for example as a \emph{credal set} in the context of imprecise probability theory \citep{wall_sr}. The concept of a credal set and related notions provide the basis of a generalization of Bayesian inference, which is especially motivated by the criticism of non-informative priors as models of ignorance in standard Bayesian inference. The basic idea is to replace a single prior distribution on $\cH$, as used in Bayesian inference, by a (credal) set of candidate priors. Given a set of observed data, Bayesian inference can then be applied to each candidate prior, thereby producing a (credal) set of posteriors. Correspondingly, any value or quantity derived from a posterior (e.g., a prediction, an expected value, etc.) is replaced by a set of such values or quantities. An important example of such an approach is robust inference for categorical data with the \emph{imprecise Dirichlet model}, which is an extension of inference with the Dirichlet distribution as a conjugate prior for the multinomial distribution \citep{bern_ai05}.

\section{Uncertainty Quantification}

According to our discussion so far, different types of learners represent their information or ``belief'' about the outcome $y_q$ for a query instance $\vec{x}_q$ in different ways:
\begin{itemize}
\item a probabilistic agent in terms of a probability distribution $q$ on $\cY$, produced by a single hypothesis $\hat{h} \in \cH$;
\item a Bayesian agent in terms of a distribution of such distributions, or a single probability distribution $q$ on $\cY$ produced on the basis of a distribution $\prob$ on $\cH$ via model averaging; 
\item a Levi agent in terms of a set $Q \subseteq \mathbb{P}(\cY)$ of distributions, induced by a set $Q' \subseteq \mathbb{P}(\cH)$ of distributions on $\cH$.
\end{itemize}
What we are mainly interested in is a quantification of the learner's epistemic and aleatoric uncertainty, i.e., we are seeking a measure of epistemic uncertainty, $\eu$, and a measure of aleatoric uncertainty, $\au$. 

For ease of notation, we subsequently omit the conditioning on the query instance $\vec{x}_q$, i.e., all probabilities of outcomes should be understood as conditional probabilities given $\vec{x}_q$ (for example, we write $\prob(y)$ instead of $\prob(y \given \vec{x}_q)$ and $\prob(y \given h)$ instead of $\prob(y \given h, \vec{x}_q)$). We denote the set of all probability distributions (probability vectors) $q = (q_1, \ldots , q_K) \in [0,1]^K$ by $\Delta_K$.


\subsection{Probabilistic Agents: Entropy}

The most well-known measure of uncertainty of a single probability distribution is the (Shannon) entropy, which, in the case of discrete $\cY$, is given as 
\begin{equation}\label{eq:shannon}
S ( q ) \defeq   - \sum_{y \in \cY} q(y) \log_2 q(y) \, ,
\end{equation}
where $0 \log 0 = 0$ by definition. This measure can be justified axiomatically, and different axiomatic systems have been proposed in the literature \citep{csis_ac08}.
It is the most obvious candidate to quantify the aleatoric uncertainty of a probabilistic agent, i.e., $\au(q) = S(q)$. As such an agent pretends to have precise knowledge of the predictive distribution, the epistemic uncertainty is 0. 



\subsection{Bayesian Agents: Entropy and Mutual Information}
\label{sec:ba2}

The Bayesian perspective, according to which the epistemic state of the learner is represented by the posterior on the hypothesis space, is quite common in the machine learning community. Most recently, the problem of predictive uncertainty estimation has attracted specific attention in the field of deep neural networks. Corresponding methods typically seek to quantify (total) uncertainty on the basis of the predictive posterior distribution on $\cY$. Moreover, epistemic uncertainty is considered as a property of the posterior $\prob( h \given \cD)$: The less concentrated this distribution is, the higher the (epistemic) uncertainty of the learner.

A principled approach to measuring and separating aleatoric and epistemic uncertainty on the basis of classical information-theoretic measures of entropy is proposed by \citet{depe_du18}. This approach is developed in the context of neural networks for regression, but the idea as such is more general and can also be applied to other settings. A similar approach was recently adopted by \citet{mobi_dc19}. 

More specifically, 
the idea is to exploit the following information-theoretic separation of the total uncertainty in a prediction, measured in terms of the (Shannon) entropy of the predictive posterior distribution (in the case of discrete $\cY$ given by (\ref{eq:shannon})): Considering the outcome as a random variable $Y$ and the hypothesis as a random variable $H$, we have
$$
S \big( Y \big) = I\big( Y , H  \big) + S \big( Y \given H \big) \, ,
$$
where $I(Y, H)$ is the mutual information between hypotheses and outcomes (i.e., the Kullback-Leibler divergence between the joint distribution of outcomes and hypotheses and the product of their marginals):
\begin{equation}\label{eq:eep}
I(Y, H)  = \mathbf{E}_{p(y,h)} \left\{  \log_2 \left( \frac{p(y,h)}{p(y) p(h)} \right) \right\} \, .
\end{equation}
This terms qualifies as a measure of epistemic uncertainty, as it captures the dependency between the probability distribution on $\cY$ and the (uncertain) hypothesis $h$. Roughly speaking, (\ref{eq:eep}) is high if the distribution $\prob(y \given h)$ varies a lot for different hypotheses $h$ with high probability. This is plausible, because the existence of different hypotheses, all considered (more or less) probable but leading to quite different predictions, can indeed be seen as a sign for high epistemic uncertainty. 

Finally, the conditional entropy is given by 
\begin{align}\label{eq:eal}
S & \big( Y \given H \big) = \mathbf{E}_{p( h \given \mathcal{D})}\left\{ S \big( \prob(y \given h) \big) \right\} = \\
&  =  - \int_{\cH} p( h \given \mathcal{D}) \left( \sum_{y \in \cY} \prob(y \given h) \log_2 \prob(y \given h) \right) \, d \, h \nonumber
\end{align}
This measure qualifies as a measure of aleatoric uncertainty: By fixing a hypothesis $h \in \cH$, the epistemic uncertainty is essentially removed. Thus, the entropy $S ( \prob(y \given h))$, i.e., the entropy of the conditional distribution on $\cY$ predicted by $h$ (for the query $\vec{x}_q$) is a natural measure of the aleatoric uncertainty. However, since $h$ is not precisely known, aleatoric uncertainty is measured in terms of the expectation of this entropy with regard to the posterior probability $p( h \given \mathcal{D})$.

\subsection{Levi Agents: Uncertainty Measures for Credal Sets}
\label{sec:umcs}

In the case of a Levi agent, uncertainty degrees ought to be specified for a set of probability distributions $Q \subseteq \Delta_K$. In the literature, such sets are also referred to as \emph{credal sets} \citep{wall_sr}, often assuming convexity (i.e., $q , q' \in Q$ implies $\alpha q + (1-\alpha) q' \in Q$ for all $\alpha \in [0,1]$). There is quite some work on defining uncertainty measures for credal sets and related representation, such as Dempster-Shafer evidence theory \citep{shaf_am}. Here, a basic distinction between two types of uncertainty contained in a credal set has been made, referred to as \emph{conflict} (randomness, discord) and \emph{non-specificity}, respectively \citep{yage_ea83}. The importance of this distinction was already emphasized by Kolmogorov \citep{kolm_ta65}. These notions are in direct correspondence with what we call aleatoric and epistemic uncertainty. The standard uncertainty measure in classical possibility theory (where uncertain information is simply represented in the form of subsets $A \subseteq \cY$ of possible alternatives) is the Hartley measure \citep{hart_to28}
\begin{equation}\label{eq:hartley}
H(A) = \log( |A|)  \, .
\end{equation}
Just like the Shannon entropy, this measure can be justified axiomatically\footnote{For example, see Chapter IX, pages 540--616, in the book by  R\'enyi \citep{reny_pt}.}. 

Given the insight that conflict and non-specificity are two different, complementary sources of uncertainty, and (\ref{eq:shannon}) and (\ref{eq:hartley}) as well-established measures of these two types of uncertainty, a natural question in the context of credal sets is to ask for a generalized representation 
\begin{equation}\label{eq:aggregate}
\on{U}(Q) = \on{AU}(Q) + \on{EU}(Q) \, ,
\end{equation}
where $\on{U}$ is a measure of total (aggregate) uncertainty, $\on{AU}$ a measure of aleatoric uncertainty (conflict, a generalization of the Shannon entropy), and $\on{EU}$ a measure of epistemic uncertainty (non-specificity, a generalization of the Hartely measure). 
Regardless of the concrete definition, a number of reasonable properties have been proposed that any uncertainty measure should obey \citep{abel_ao05,pan_an19}.
%

As for the non-specificity part in (\ref{eq:aggregate}), the following generalization of the Hartley measure to the case of graded possibilities has been proposed by various authors \citep{abel_an00}:
\begin{equation}\label{eq:gh}
\on{GH}(Q) \defeq  \sum_{A \subseteq \cY} \on{m}_Q(A) \, \log(|A|) \, ,
\end{equation}
where $\on{m}_Q: \,  2^{\cY} \longrightarrow [0,1]$ is the M\"obius inverse of the capacity function $\nu :\, 2^{\cY} \longrightarrow [0,1]$ defined by
\begin{equation}\label{eq:cap}
\nu_Q(A) \defeq \inf_{q \in Q} q(A) 
\end{equation}
for all $A \subseteq \cY$, that is,
$$
\on{m}_Q(A) = \sum_{B \subseteq A} (-1)^{|A \setminus B|} \nu_Q(B) \, .
$$
This measure is ``well-justified'' in the sense of possessing a sound axiomatic basis and obeying a number of desirable properties \citep{klir_ot87}. 

Interestingly, an extension of Shannon entropy, ``well-justified'' in the same sense, has not been found so far. 
As a possible way out, it was then suggested to define a meaningful measure of total or aggregate uncertainty, and to \emph{derive} a generalized measure of aleatoric uncertainty via \emph{disaggregation}, i.e., in terms of the difference between this measure and the measure of epistemic uncertainty (Hartley), or vice versa, to derive a measure of epistemic uncertainty as the difference between total uncertainty and a meaningful measure of aleatoric uncertainty. 
The upper and lower Shannon entropy play an important role in this regard: 
\begin{equation}\label{eq:gg}
S^*(Q) \defeq \max_{q \in Q} S(q) \, , \quad
S_*(Q) \defeq \min_{q \in Q} S(q)
\end{equation}
Based on these measures, the following disaggregations of total uncertainty (\ref{eq:aggregate}) have been proposed \citep{abel_dt06}:
\begin{align}
S^*(Q) & = \big(S^*(Q) - \on{GH}(Q) \big)  + \on{GH}(Q)  \label{eq:unc1} \\
S^*(Q) & = S_*(Q)  + \big(S^*(Q) - S_*(Q) \big)   \label{eq:unc2}
\end{align}
In both cases, upper entropy serves as a measure of total uncertainty $U(Q)$, which is again justified on an axiomatic basis. In the first case, the generalized Hartley measure is used for quantifying epistemic uncertainty, and aleatoric uncertainty is obtained as the difference between total and epistemic uncertainty. In the second case, lower entropy is used as a (well-justified) measure of aleatoric uncertainty, and epistemic uncertainty is derived in terms of the difference between upper and lower entropy.

\section{Ensemble-Based Uncertainty Quantification}


Ensemble-based approaches to uncertainty quantification have recently been advocated by several authors. Intuitively, ensemble methods should indeed be suitable to represent information about epistemic uncertainty. This information should essentially be reflected by the variability of the individual ensemble members: If all ensemble members are very similar to each other, in the sense of producing similar predictions, the learner is apparently not very uncertain about the best hypothesis. On the other side, a high variability among ensemble members suggests a high degree of (epistemic) uncertainty.

Adopting a Bayesian perspective, the variance of the predictions produced by an ensemble is inversely related to the ``peakedness'' of a posterior distribution $\prob(h \given \mathcal{D})$. Thus, an ensemble can be considered as an approximate representation of a second-order distribution $p( h \given \mathcal{D})$ in a Bayesian setting. An interesting special case is the (implicit) construction of ensembles using techniques like Dropout \citep{gal_bc16} or DropConnect \citep{mobi_dc19} in deep neural networks \citep{hara_ao16}.

 
Uncertainty quantification based on ensemble techniques is also advocated by \citet{laks_sa17}, who propose a simple ensemble approach as an alternative to Bayesian NNs. The authors note that Monte Carlo dropout (MC-dropout) can be interpreted as an approximate Bayesian approach and as an ensemble technique at the same time. According to their opinion, ``the ensemble interpretation seems more plausible particularly in the scenario where the dropout rates are not tuned based on the training data, since any sensible approximation to the true Bayesian posterior distribution has to depend on the training data.'' They take this as a strong motivation for the independent investigation of ensembles as an alternative solution for estimating predictive uncertainty. 

Given this motivation, we address the question of how the measures of uncertainty introduced above can be realized by means of ensemble techniques, i.e., how they can be computed (approximately) on the basis of a finite ensemble of hypotheses $H = \{ h_1, \ldots , h_M \}$, which can be thought of as a sample from the posterior distribution $p( h \given \mathcal{D})$. More specifically, we consider this question for the case of a Bayesian and a Levi agent. The following notation will be used: 
\begin{itemize}
\item $p_{k,m} = \prob(y_k \given h_m, \vec{x}_q)$ is the probability predicted for class $y_k$ by hypothesis $h_m$ for query $\vec{x}_q$, i.e., $(p_{1,m}, \ldots , p_{K,m}) = \prob( \cdot \given h_m, \vec{x}_q)$;
\item $l_m = \prob( \mathcal{D} \given h_m)$ denotes the likelihood of $h_m$;
\item $q_k =  \sum_{m=1}^M \prob( h_m \given \mathcal{D}) \, p_{k,m}$ is the posterior probability estimate for class $y_k$ produced by the ensemble through weighted averaging.

\end{itemize}
 
\subsection{Bayesian Agents}

Recalling the approach presented in Section \ref{sec:ba2}, it is obvious that (\ref{eq:eep}) and (\ref{eq:eal}) cannot be computed efficiently, because they involve an integration over the hypothesis space $\cH$. Based on an ensemble $H = \{ h_1, \ldots , h_M \}$, an approximation of (\ref{eq:eal}) can be obtained by
\begin{equation}\label{eq:approxua}
\au(\vec{x}_q) \defeq
-  \sum_{m=1}^M  \prob( h_m \given \mathcal{D}) 
\sum_{k=1}^K p_{k,m} \log_2 p_{k,m}   \, ,
\end{equation}
an approximation of total uncertainty, i.e., Shannon entropy (\ref{eq:shannon}), by
\begin{equation}\label{eq:approxut}
\tu(\vec{x}_q) \defeq  - \sum_{k=1}^K q_k  \log_2 q_k \, ,  
\end{equation} 
and finally an approximation of (\ref{eq:eep}) by $\eu(\vec{x}_q) = \tu(\vec{x}_q) - \au(\vec{x}_q)$.
Assuming a uniform prior, which is quite natural in the case of ensembles, the posterior probability of hypotheses can be obtained from $\prob( h_m \given \mathcal{D}) \propto l_m$.

\subsection{Levi Agents}
\label{sec:leviensemble}
How could the idea of a Levi agent be implemented on the basis of an ensemble approach? As explained above, credal inference yields a set of probability estimates, each of which is obtained by Bayesian model averaging according to a different prior. Thus, instead of assuming a uniform prior $\prob( h_m) \equiv 1/M$, we should now proceed from a set of priors. A simple example is the family 
\begin{equation}\label{eq:sdelta}
S_\delta = \left\{ \vec{s} = ( s_1, \ldots , s_M) \given \frac{1}{\delta \, M}  \leq s_m \leq \frac{\delta}{M}  , \, 
\sum_{m=1}^M s_m = 1 \right\} 
\end{equation}
of distributions $\delta$-close to uniform, where $\delta \geq 1$ is a (hyper-)parameter. Thus, compared to the uniform prior, the probability of a single hypothesis can now be decreased or increased by a factor of at most $\delta$. The set of posterior probabilities is then given by 
$$
\left\{ 
\prob( h_m \given \mathcal{D}) = \frac{s_m \, l_m}{\sum_{i=1}^M  s_i \, l_i}  \given  \vec{s} \in S_\delta 
\right\} \, ,
$$
and hence the credal set on $\cY$ by
$$
Q = \left\{ q = \sum_{m=1}^M s_m l_m h_m(\vec{x}_q) / \sum_{m=1}^M s_m l_m \given \vec{s} \in S_\delta
\right\}
$$
To compute the decompositions (\ref{eq:unc1}) and  (\ref{eq:unc2}) for $Q$, we need to compute the measures $S^*$, $S_*$, $\on{GH}$. 
According to (\ref{eq:gh}), the computation of the measure $\on{GH}$ requires the capacity (\ref{eq:cap}), i.e., the lower probability $\nu_{Q}(A)$ of each subset of classes $A \subseteq \cY$. For $A= \{ y_j \}_{j \in J}$ identified by an index set $J \subseteq [K]$, the latter is given by
$$
\nu_Q(A) = \min_{q \in Q} q(A) = \min_{\vec{s} \in S_\delta}  \frac{\sum_{j \in J} \sum_{m=1}^M s_m\, l_m\, p_{j,m}}{\sum_{m=1}^M s_m \, l_m} \, .
$$
Thus, finding $\nu_Q(A)$ comes down to solving a linear-fractional programming problem (for which standard solvers can be used). Moreover, finding $S^*$ comes down to solving 
\begin{align*}
& \max_{\vec{s} \in S_\delta} \sum_{k=1}^K 
\frac{\sum_{m=1}^M s_m \, l_m \, p_{k,m}}{\sum_{m=1}^M s_m \, l_m} 
\log \frac{\sum_{m=1}^M s_m \, l_m \, p_{k,m}}{\sum_{m=1}^M s_m \, l_m} \, ,
\end{align*}
and similarly for $S_*$ (with $\max$ replaced by $\min$).

\section{Experiments}\label{sec:exp}

Predicted uncertainties are often evaluated indirectly, for example by assessing their usefulness for improved prediction and decision making, because  the data does normally not contain information about any sort of ``ground truth'' uncertainties. Here, we conducted such an evaluation by producing \emph{accuracy-rejection curves}, which depict the accuracy of a predictor as a function of the percentage of rejections \citep{mpub170}: A learner, which is allowed to abstain on a certain percentage $p$ of predictions, will predict on those $(1-p)$\,\% on which it feels most certain. Being able to quantify its own uncertainty well, it should improve its accuracy with increasing $p$, hence the accuracy-rejection curve should be monotone increasing (unlike a flat curve obtained for random abstention).

\subsection{Data Sets and Experimental Setting}
\label{sec:exLocal}

We compare the Bayesian agent with different variants of the Levi agent in terms of their ability to quantify aleatoric and epistemic uncertainty. The Bayesian agent quantifies these uncertainties according to (\ref{eq:approxua}) and (\ref{eq:approxut}).
The Levi agent is implemented as described in Section \ref{sec:leviensemble}. Uncertainty is quantified based on the generalized Hartley measure (Levi-GH) according to (\ref{eq:unc1}), or based on upper and lower entropy (Levi-Ent) according to (\ref{eq:unc2}). In this experiment, we set the hyper-parameter $\delta=2$.

We performed experiments on various well-known data sets from the UCI repository\footnote{\url{http://archive.ics.uci.edu/ml/index.php}}.
The data sets are randomly split into 70\% for training and 30\% for testing, and accuracy-rejection curves are produced on the latter. Each experiment is repeated and averaged over 100 runs.
We create ensembles using the Random Forest Classifier from SKlearn. The number of trees within the ensemble is set to 10. Each tree can grow to a maximum of 10 splits. Probabilities are estimated by (Laplace-corrected) relative frequencies in the leaf nodes of a tree.

\subsection{Results}

\begin{figure}
\begin{center}

  \subfloat{\includegraphics[scale=0.4]{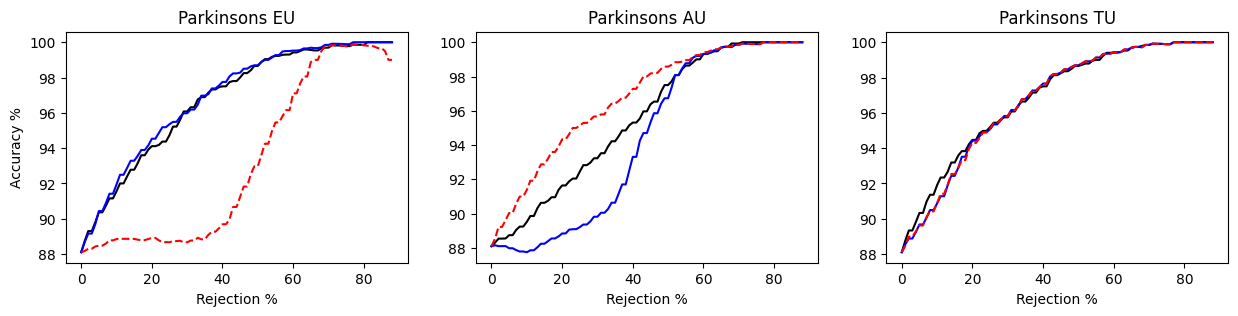} }\\
  \subfloat{\includegraphics[scale=0.4]{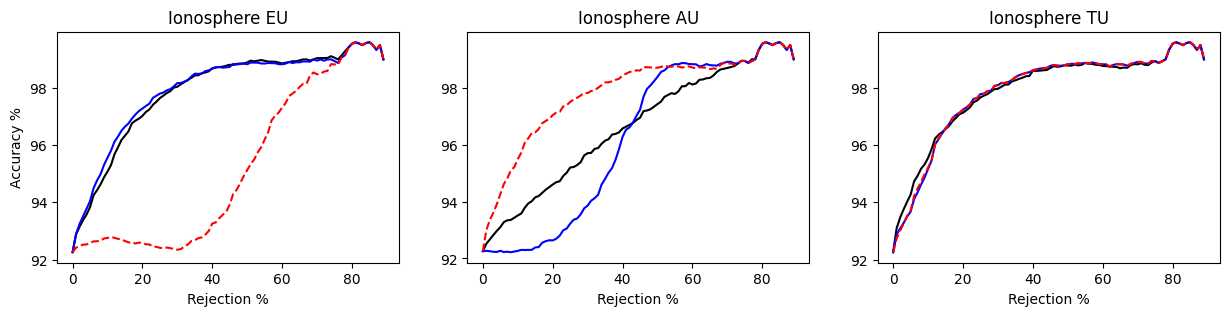} }\\
  \subfloat{\includegraphics[scale=0.4]{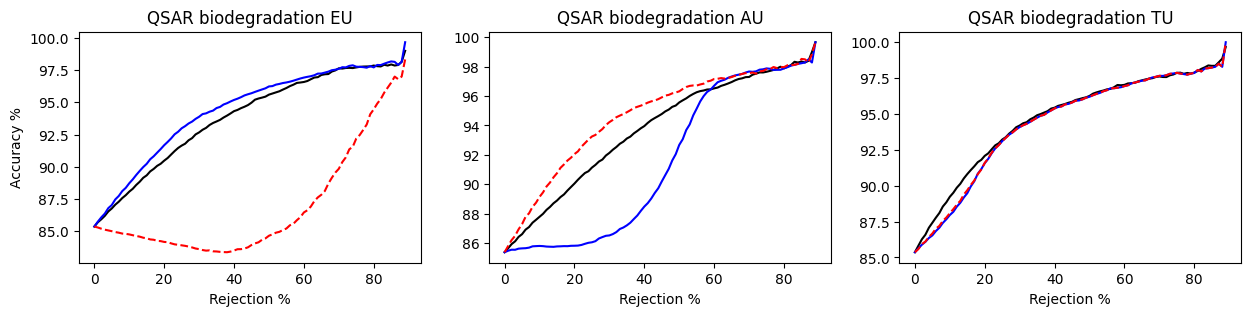} }\\
  \subfloat{\includegraphics[scale=0.4]{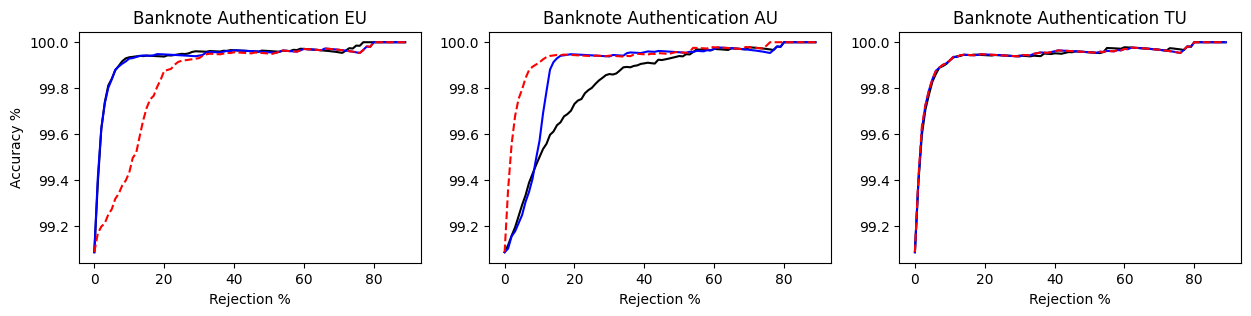} }\\
  \subfloat{\includegraphics[scale=0.4]{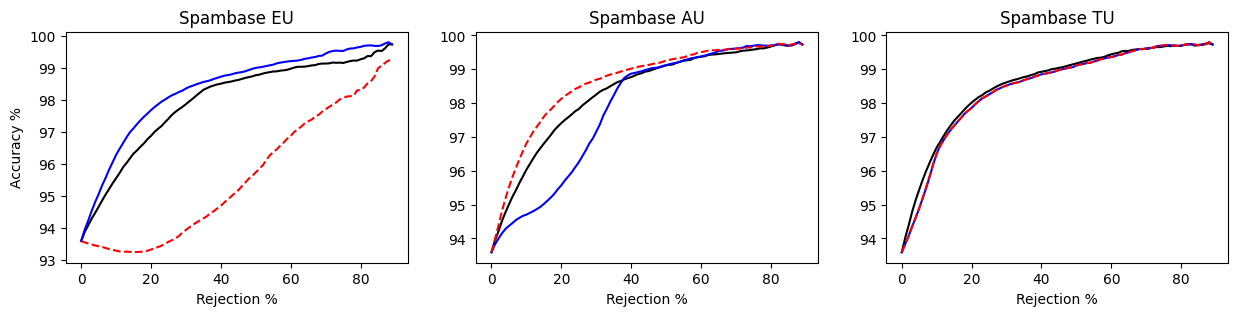} }\\

  \subfloat{\includegraphics[scale=0.25]{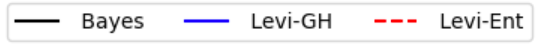} }

\caption{Accuracy-rejection curves for the Bayesian and the Levi agent.}
\label{fig:unc1}
\end{center}
\end{figure}

Fig.\ \ref{fig:unc1} shows the accuracy-rejection curves for the different learners, separated into epistemic uncertainty (EU) in the left, aleatoric uncertainty (AU) in the middle, and total uncertainty (TU) on the right column. Due to space restrictions, we only show the results for five data sets, noting that the results for other data sets are very similar. The following observations can be made.
\begin{itemize}
\item
As suggested by the shape of the accuracy-rejection curves, both the Bayesian and the Levi agent perform quite well in general. On total uncertainty, they are basically indistinguishable, which is almost a bit surprising, given that these uncertainties are quantified on the basis of different principles.
\item
Levi-GH seems to have an advantage over the Bayesian agent on epistemic uncertainty, providing evidence for the generalized Hartley measure as a reasonable measure of epistemic uncertainty. 

\item
Levi-Ent seems to have an advantage over the Bayesian agent on aleatoric uncertainty, providing evidence for the lower entropy as a reasonable measure of aleatoric uncertainty. 

\item
The ``derived'' measures, $S^*(Q) - \on{GH}(Q)$ for aleatoric and $S^*(Q) - S_*(Q)$ for epistemic uncertainty, both perform quite poorly.


\end{itemize}

\section{Conclusion}

We proposed a distinction between different types of uncertainty-aware learning algorithms, discussed measures of total, aleatoric and epistemic uncertainty of such learners, and developed ensemble-methods for approximating these measures. In particular, we compared the classical Bayesian approach with what we call a Levi agent, which makes predictions in terms of credal sets. 

In an experimental study on uncertainty-based abstention, both methods show strong performance. While the Bayesian and the Levi agent are on a par for total uncertainty, improvements of the Bayesian approach can be achieved for the two types of uncertainty separately: The generalized Hartley measure appears to be superior for epistemic and the lower entropy for aleatoric uncertainty quantification. On the other side, the alternative measures of aleatoric and epistemic uncertainty obtained through disaggregation perform quite poorly. These results can be seen as an interesting empirical complement to the theoretical (axiomatic) research on uncertainty measures for credal sets.  

In future work, we seek to further deepen our understanding of ensemble-based uncertainty quantification and elaborate on the approach presented in this paper. An interesting problem, for example, is the tuning of the (hyper-)parameter $\delta$ in (\ref{eq:sdelta}), for which we simply took a default value in the experiments. Obviously, this parameter has an important influence on the uncertainty of the Levi agent. Besides, we also plan to develop alternative approaches for constructing ensembles. Last but not least, going beyond abstention and accuracy-rejection curves, we plan to apply and analyze corresponding methods in the context of other types of uncertainty-aware decision problems.

\subsubsection*{Acknowledgments}
This work has been supported by the German Federal Ministry of Education and Research (BMBF) within the project EML4U under the grant no 01IS19080 A and B.



\end{document}